\documentclass[lettersize,journal]{IEEEtran}
\usepackage{amsmath,amsfonts}
\usepackage{algorithmic}
\usepackage{array}
\usepackage[caption=false,font=normalsize,labelfont=sf,textfont=sf]{subfig}
\usepackage{textcomp}
\usepackage{stfloats}
\usepackage{url}
\usepackage{verbatim}
\usepackage{graphicx}
\def\BibTeX{{\rm B\kern-.05em{\sc i\kern-.025em b}\kern-.08em
		T\kern-.1667em\lower.7ex\hbox{E}\kern-.125emX}}
\usepackage{cite}
\usepackage{balance}
\usepackage{siunitx}
\usepackage{threeparttable}
\usepackage{tabularx}
\usepackage{multicol}
\usepackage{multirow}
\usepackage{xfrac}
\usepackage{soul} 		
\usepackage{xcolor} 	

\begin{document}

\title{A Quantitative Autonomy Quantification Framework for Fully Autonomous Robotic Systems}
\author{Nasser Gyagenda, Hubert Roth
\thanks{N. Gyagenda and H. Roth are with the Department of Electrical Engineering and Computer Science of the University of Siegen, Siegen, Germany. \\Email: nasser.gyagenda@uni-siegen.de, hubert.roth@uni-siegen.de}
}

\maketitle
\begin{abstract}
Although autonomous functioning facilitates deployment of robotic systems in domains that admit limited human oversight on our planet and beyond, finding correspondence between task requirements and autonomous capability is still an open challenge. Consequently, a number of methods for quantifying autonomy have been proposed over the last three decades, but to our knowledge all these have no discernment of sub-mode features of variation of autonomy and some are based on metrics that violet the Goodhart’s law. This paper focuses on the full autonomous mode and proposes a quantitative autonomy assessment framework based on task requirements. The framework starts by establishing robot task characteristics from which three autonomy metrics, namely requisite capability set, reliability and responsiveness are derived. These characteristics were founded on the realization that robots ultimately replace human skilled workers, from which a relationship between human job and robot task characteristics was established. Additionally, mathematical functions mapping metrics to autonomy as a two-part measure, namely of level and degree of autonomy are also presented. The distinction between level and degree of autonomy stemmed from the acknowledgment that autonomy is not just a question of existence, but also one of performance of requisite capability. The framework has been demonstrated on two case studies, namely autonomous vehicle at an on-road dynamic driving task and the DARPA subterranean challenge rules analysis. The framework provides not only a tool for quantifying autonomy, but also a regulatory interface and common language for autonomous systems developers and users. Its greatest feature is the ability to monitor system integrity when implemented online.
\end{abstract}


\section{Introduction}
\label{sect:introduction}

\IEEEPARstart{H}{umans} have been the main higher-level controllers for most dynamic systems due to their perceptual, manipulation, decision-making, problem-solving and other cognitive and physical abilities. But as robotic systems find applications in more complex and highly dynamic environments whose control demands exceed human regulation capability, system stability under exclusive human control in an open world cannot be guaranteed. Furthermore, human control integrity fluctuates, leading to poor decisions and slow responses that sometimes result in accidents. This phenomenon is known as the human error and is responsible for \(94\pm2.2\%\) of road vehicle crashes \cite{singh2015critical} and \(80\%\) of civil aviation accidents \cite{rankin2007meda}. This calls for the adoption of machine-based control. But such a paradigm shift requires machines to have the ability to make at least one of three deliberate choices at their own volition guided by their belief, namely (1) choice of a goal; (2) choice of course of action for achieving a goal; (3) choice of goal-action pair. Such systems are known as autonomous systems. This led to the definition of autonomy as the system's ability to select an intermediate goal and/or course of action for achieving that goal, as well as approve or disapprove any previous and future choices while achieving its overall goal.

When operating autonomously, systems should act independent of external intervention in an acceptable manner. This begs the question of how to design a consistent and reliable policy for acceptability. To quote H. James Harrington, “Measurement is the first step that leads to control and eventually to improvement." Following this, it is not only necessary to characterize tasks, but also to establish metrics that are easily measurable, sensitive to differences in performance, extensive enough to capture autonomy evolution in a system, admit quantitative analysis and with good output resolution.

This work proposes a set of quantitative autonomy metrics derived from robot task characteristics, which were in turn inferred from relating robot task characteristics to human job characteristics applied in industries. We realized that autonomy is purposive and domain/performance-specific, where the former means that autonomous functions are designed to address a specific goal, while the latter highlights the fact that autonomous functioning is associated with predefined performance requirements. Therefore, autonomy for a system is defined with respect to a purpose and performance. Since performance requirements change as new environments and applications emerge, a good autonomy assessment framework must be sensitive to these changes. Such a framework based on system capability is presented herein and its details described in the following sections: Sect.~\ref{sect_background} puts this work in context with the existing related literature; Sect.~\ref{sect_autonomyMeasure} gives a detailed description of the proposed metrics, their associated mathematical definitions, and level of autonomy and degree of autonomy models; Sect.~\ref{sect_integrity} describes the integrity monitoring process integrated into the framework; Sect.~\ref{sect_caseStudy} provides application demonstrations for the proposed autonomy framework; Sect.~\ref{sect_conclusion} presents the concluding remarks.

The framework supports not only quantitative quantification of autonomy, but also integrity monitoring through online implementation. The former affords mathematical analysis, while the latter is key in ensuring safety in complex dynamic systems. Furthermore, with interval and ratio-scale measures, and establishment of quantitative essential and functional performance requirements, it is expected to boost human trust in autonomous systems as well as simplify the processes of their design, development and regulation.

\section{Background and Related Work}
\label{sect_background}

\noindent Autonomous robots are increasingly sought after in a number of applications for their benefits, among which include replacing human operators \cite{clough2002metrics}, safety risk assessment, handling of complex applications and attaining time critical responses \cite{jackson1991achieving},  operating domain qualification and delineation \cite{lampe2006performance}, basis for control architecture selection \cite{insaurralde2012autonomy}, characterization of autonomous technologies and assessing technology maturity \cite{kendoul2013towards}, operator training and licensing \cite{clothier2014review}, and reducing operating costs whilst maximizing mission success \cite{elbanhawi2017enabling} among others. These have motivated a global surge in the production and marketing of autonomous systems, which raises regulatory and ethical concerns. Although proper regulatory laws and good engineering practices can address these concerns, their formulation requires autonomy quantification tools and metrics. \par

Over the years, a number of autonomy assessment frameworks have been proposed. These frameworks are founded on the researchers' definition of autonomy, which influenced their choice of factors that characterize autonomy of systems. This section reports on such frameworks in the existing literature found relevant to this research work.  \par

One of the earliest works on autonomy is \cite{jackson1991achieving}, where autonomy is defined as a measure of supervision. They assert that the less external commands the system requires during operation and the less data the system sends to an external controller for interpretation, the more control processes it is capable of handling onboard, hence, the higher its autonomy. This bandwidth measure assumes commands to have equal importance to the fulfilment of the task, which is generally true only for very simple tasks. Additionally, the entropy of each command and the knowledge representation may differ resulting in varying amounts of information per command. \par

ALFUS workgroup of NIST (National Institute of Standards and Technology) defines autonomy as a system's ability to use its sensing, perception, analysis, planning, decision-making, communication and actuation capabilities to achieve an assigned goal \cite{huang2007autonomy}. Their proposed assessment framework differentiates between autonomy and level of autonomy, where the former is a function of task complexity, degree of operator independence and environment complexity, while the latter is a measure of the degree of operator independence. But it provides neither task nor environment characterizations, and the applied metrics lack quantitative interpretations needed in order to support mathematical analysis. \par

Autonomy has been likened to intelligence and performance, though related, the three terms differ. Intelligence is the ability of a system to acquire and/or use knowledge, while performance is a functional quality of a system. Intelligence and performance differ in the way they relate with autonomy. To contextualize this, autonomy can be fulfilled by artificial intelligence methods or deterministic processes. By observation, both implementations are considered intelligent, extrinsic intelligence or intelligent by observation\textemdash exercising abilities considered intelligent when performed by humans \cite{russell2010artificial}. In actuality, only the former is intelligent by design, intrinsic intelligence\textemdash a system built from intelligent building blocks. Extrinsic intelligence is incognisant of the underlying implementation, rendering it incapable of distinguishing automatic from autonomous functioning, and is not considered herein. An example of this is \cite{lampe2006performance}, where autonomy is defined as a measure of performance with two metrics, namely environment complexity and information quality. As noted before, information quality is not only a bad metric, but also oblivious to performance differences between different implementations of any particular full autonomous capability. \par

Autonomy assessment frameworks can be categorised based on their output format into ordinal, ratio and interval levels of measurements \cite{clothier2014review}. The ordinal-scale frameworks divide autonomy into discrete levels, as in the ACL (Autonomy Control Level) chart in \cite{clough2002metrics}. The chart has eleven levels ranging from remotely operated systems at level zero to fully autonomous systems at level eleven, and is based on OODA-loop inspired metrics including perception, analysis, decision making and capability. A similar eleven level chart was proposed in \cite{huang2007autonomy}, but based on operator independence. They then extended the non-contextual autonomy level measure to a ratio-scale contextual measure that was based on autonomy level, task complexity and environment complexity.\par

A commonly referenced autonomy level chart is the SAE’s levels of driving automation \cite{sae2018surface}. Following this chart, a manufacturer assigns a vehicle to one of six levels based on human involvement in the control process. Such an approach is neither able to discern performance differences between vehicles at a similar level nor quantify contextual performance. Consequently, the department of motor vehicles of the state of California devised the disengagement rate metric \cite{CaliforniaDMV2022}. But this online metric is incapable of capturing performance nuances of a driving system, i.e., it does not account for environment complexity, integrity, availability or driving quality. Worst of all, one can strategically select test routes, time of day, duration and/or weather conditions to minimize disengagement rate, rendering it a bad measure according to Goodhart’s law. \par

A batch test of frameworks \cite{clough2002metrics,kendoul2013towards,huang2007autonomy} and seven others on classification of six UASs (Unmanned Aircraft Systems) in \cite{clothier2014review} revealed that only four frameworks were able to classify all six vehicles, but only one was able to unambiguously classify all six. Unfortunately, even the one that unambiguously classified all six, failed to distinguish between a vehicle with unsupervised capabilities, a vehicle with supervised capabilities and a purely remotely operated vehicle. \par

Similar to \cite{huang2007autonomy}, the framework in \cite{insaurralde2012autonomy} outputs a ratio-scale measure. This framework is based on the premise that autonomous systems have an automated problem-solving process, so it measures autonomy as levels of automation of the problem-solving process using either Eq.\ref{eqn_doa_ins1} or Eq.\ref{eqn_doa_ins2}. As such, this limits autonomy to deterministic implementations. Other works with a ratio-scale autonomy measure include \cite{doboli2018cities} and \cite{curtin2005autonomous}, where autonomy is given by Eq.~\ref{eqn_doa_dob} and Eq.~\ref{eqn_doa_cur} respectively. \par

\begin{equation}\label{eqn_doa_ins1}
	\alpha = \frac{\delta_D + \delta_E + \delta_S + \delta_I + \delta_V}{n}
\end{equation} where \(n\) is the number of autonomous capabilities considered and \(\delta_D\), \(\delta_E\), \(\delta_S\), \(\delta_I\) and \(\delta_V\) are levels of automation for definition, exploration, selection, implementation and verification capabilities respectively.

\begin{equation} \label{eqn_doa_ins2}
	\alpha = 10\cdot\frac{\frac{\eta_{D,act}}{\eta_{D,std}}+\frac{\eta_{E,act}}{\eta_{E,std}}+\frac{\eta_{S,act}}{\eta_{S,std}}+\frac{\eta_{I,act}}{\eta_{I,std}}+\frac{\eta_{V,act}}{\eta_{V,std}}}{n}
\end{equation} where \(\eta\) quotients are ratios of actual (act) to standard (std) system behaviour for definition (D), exploration (E), selection(S), implementation (I) and verification (V) capabilities.

\begin{equation} \label{eqn_doa_dob}
	\alpha = \int_{T_i}^{T_f}\int_{V_i}^{V_f}\int_{PU}^{PL}F \,dp\,da\,dt
\end{equation} where \(F\), \(p\), \(a\) and \(t\) are human effort, performance, area and time respectively.

\begin{equation} \label{eqn_doa_cur}
	\alpha = C_n\left(\frac{B_C}{B_T}\right)^{-i}\left(\frac{T_C}{T_T}\right)^{-j}
\end{equation} where \(B_C\), \(B_T\), \(T_C\) and \(T_T\) are control bits, total message size, contact time and total mission time respectively, while \(C_n\), \(i\) and \(j\) are constants.

Unlike the above reported approaches, which output either an ordinal or ratio-scale measure, the one presented herein outputs a hybrid measure consisting of an ordinal-scale measure known as LoA (Level of Autonomy) and its associated ratio-scale measure known as DoA (Degree of Autonomy) for fully autonomous systems. This hybridization resulted from the realization that autonomous systems differ not only in the existent capabilities (kind), but also in capability performance (degree). To assess LoA, the framework adopted the autonomy level chart structures of \cite{huang2007autonomy} and \cite{clough2002metrics}, but with custom level descriptions. More details on LoA and DoA are given in Sect.~\ref{sect_LoADoA}. In establishing autonomy metrics, works like \cite{hasslacher1995living} applied biomorphic survival laws, we derived ours from human job characteristics to reflect the purposiveness of robotic systems. This makes sense since robots ultimately replace human skilled workers. The proposed metrics and their relation to LoA and DoA are described in the next section. \par

\section{Autonomy Quantification}
\label{sect_autonomyMeasure}

\noindent Quantifying autonomy requires establishment of appropriate metrics. But how does one identify such metrics? The search thereof should focus on an impartial search space, steer away from particular biases and distinctive system features. Following these principles, the resulting metrics should show invariance to implementation choices such as of state estimators and sensor fusion methods, software architecture, computing architecture and modelling strategy to mention a few. \par

As autonomous systems ultimately replace human skilled workers, it was prudent to start by examining human-job characteristics applied in industries. This examination identified the four characteristics indicated in the left column of Table~\ref{tab:humanJobXtics}, while in the right column are their translations to robot-task characteristics. This constrained the metrics search to these four characteristics. In the review on autonomy assessment criteria in \cite{insaurralde2012autonomy}, it is indicated that \(25\%\) of the considered twenty frameworks viewed autonomy from environment, mission and self-autonomy perspectives. This pointed to the proposed characteristics having merit, as well as the obliviousness of their relevance to majority of the previous research works. \par

\renewcommand{\arraystretch}{1.3}
\linespread{1}
\begin{table}
	\begin{center}
	\caption{Human-job characteristics mapped to robot-task characteristics. The former and their associated importance weights were adopted from \cite{lytle1946job}.}
	\label{tab:humanJobXtics}
	\begin{tabularx}{0.94\linewidth}{X|c|X}\hline
		\textbf{Human-Job \newline Characteristics} & \textbf{Importance} & \textbf{Robot-Task \newline Characteristics} \\ \hline
		Skills & \(50\%\) & Capability \\ \hline
		Responsibility & \(25\%\) & Reliability \\ \hline
		Effort & \(15\%\) & Responsiveness \\ \hline
		Working conditions & \(10\%\) & Environment complexity \\ \hline \hline
		\hspace*{\fill}\textbf{Total} & \textbf{100\%} & \\ \hline
	\end{tabularx}
	\end{center}
\end{table}

Although a skilled human was taken as a reference in identifying the robot-task characteristics, the goal of developing autonomous robotic systems with as diverse a set of capabilities as that of a skilled human’s is one considered unnecessary as these systems are purposive and domain specific. Therefore, herein, autonomous functioning is limited to a specific task and operating domain.

\subsection{Autonomy Metrics}
\label{subsect_autonomyMetrics}

\noindent The desiderata for solving any task is possession of a requisite capability set that meets the performance requirements of that task and operating domain. Therefore, the sought after metrics should capture both the \textit{existence} of essential capabilities as well as their \textit{performance}. It follows that autonomy as measured by degree of supervision \cite{jackson1991achieving} is just one part of the whole story. In the search for metrics, as it is for end users and regulatory authorities, it is not technology, but capability and performance thereof that are crucial.\par

A capability exists and therefore functional if its performance as measured by its accuracy and speed of response is acceptable. But since the true value is normally not known, it is not always possible to determine the actual error, and hence, accuracy. Instead, the statistical bound on error, variance is adopted. On observing that reliability and responsiveness are functions of variance and rate of response respectively, they were selected as two of the metrics. The acceptance policy is conditioned on compliance with the essential performance requirements, which are set in accordance with the law of requisite variety to account for both task and operating domain characteristics. Therefore, the environment complexity characteristic is subsumed by reliability and responsiveness through their established performance requirements. \par

Two types of requirements have been identified, namely essential and functional requirements. The former guarantees non-contextual autonomous performance, also known as passive or offline autonomy \cite{insaurralde2012autonomy}, while the latter guarantees contextual autonomous performance, also known as active \cite{insaurralde2012autonomy} or online autonomy and is key to ensuring system integrity as described in Sect.~\ref{sect_integrity}. With the requirements promulgated, the developers then design and implement control strategies for their achievement on any preferred technology. In summary, the three proposed metrics are requisite capability set, reliability and responsiveness. \par

\subsection*{Capability}

\noindent Robotic systems are built for a particular task or to demonstrate a particular emergent behaviour. With the assumption of a behaviour-based paradigm, solving a task becomes a matter of systematic coordination of behaviours. Therefore, one can rightly state that underlying any task-solving ability is a set of behaviours and a behaviour-coordination mechanism. Herein we refer to a coordinated sequence of behaviours as a capability. Although capabilities may share underlying behaviours, each is considered independent. \par

As indicated in Table~\ref{tab:humanJobXtics}, capabilities are the most important of all robot-task characteristics. Also a review of twenty autonomy assessment frameworks in \cite{insaurralde2012autonomy} revealed capability study as the dominant approach followed by mission, interaction and environment, to mention the top four. From the same study, the dominant capabilities included problem solving, motion control, perception, communication, acquisition and self-preservation. A comprehensive list includes localization, motion control, trajectory/path following, object detection, object recognition, object tracking, motion planning, communication, object manipulation, integrity monitoring and mapping. \par

The concept of associating each human-job with a set of skills is extendable to robot tasks through association of each robot-task with a set of capabilities. The associated set then becomes a tool for scoring the capability aspect of the system. As indicated in Table~\ref{tab_applicationCapabilities_1}, different applications require different sets of capabilities. Each set is divided into requisite capabilities, which represent a minimal mandatory subset for achieving a particular task, and auxiliary capabilities, which extend the set to improve on its versatility and robustness. A system may possess capabilities beyond these two categories for supportive purposes. When assessing autonomy of a system at a particular task, however, only the requisite capability subset is considered for evaluation, as it constitutes the minimal set required for completing that task, and hence, necessary for autonomous functioning. The task execution strategy follows a capability coordination plan generated by a higher-level planner.

Execution of any requisite capability constitutes a hazard, which has to be properly managed. The risk management process identifies the risk, assesses its severity and provides the necessary control measures. The risk control measure adopted herein is establishment of performance limits (within which the risk is acceptable), by manufacturers and regulatory authorities in accordance with their risk acceptance policy and standards respectively. Acceptance is ascertained through compliance with functional requirements, i.e., reliability and responsiveness requirements.

\renewcommand{\arraystretch}{1.3}
\linespread{1}
\begin{table*}
	\begin{center}
	\caption{Autonomous mobile robotic applications and their capabilities.}
	\label{tab_applicationCapabilities_1}
	\begin{tabularx}{\linewidth} {
			>{\raggedright\arraybackslash}p{5.4cm}
			| >{\raggedright\arraybackslash}X
			| >{\raggedright\arraybackslash}p{3.3cm} }\hline
		\textbf{Application} & \textbf{Requisite Capability} & \textbf{Auxiliary Capability} \\ \hline
		Search and rescue with: a single vehicle; multiple vehicles.
		& Localization, motion control, path following/trajectory tracking, object detection, motion planning, and communication for multiple vehicles & Inter-vehicle coordination, object recognition,
		\\ \hline
		Emergency response to: natural disasters; distress calls.
		& Localization, motion control, path following/trajectory tracking, communication, motion planning, object detection & Inter-vehicle coordination, object recognition,
		\\ \hline
		Remote sensing for: environmental monitoring; meteorology.
		& Localization, motion control, path/trajectory tracking, communication, motion planning, object detection & Inter-vehicle coordination, object recognition,
		\\ \hline
		Delivery service for: healthcare supplies; logistics; disaster relief.
		& Localization, motion control, path following/trajectory tracking, motion planning & Communication, object detection, object recognition  \\ \hline
		Wireless communication network with: a single vehicle; multiple vehicles.
		& Localization, motion control, path/trajectory tracking, motion planning, and communication for multiple vehicles & Inter-vehicle coordination for multiple vehicles  \\ \hline
		Inspection and surveillance of: coast lines; construction sites; infrastructure; irrigation channels; property; assets.
		& Localization, motion control, path following/trajectory tracking, communication, motion planning, object detection & Object recognition, object tracking  \\ \hline
		Real-time disaster monitoring for: forest fires; landslides; floods; avalanche; volcanic eruptions.
		& Localization, motion control, path following/trajectory tracking, communication, motion planning, object detection & Object recognition, object tracking
		\\ \hline
		Photography and videography in: filming; journalism; photogrammetry; real-estate.
		& Localization, motion control, path following/trajectory tracking, communication, motion planning, object detection, object tracking & Object recognition  \\ \hline
		Mapping and survey for: archaeological documentation; urban planning; land use mapping; mining.
		& Localization, motion control, path following/trajectory tracking, mapping, communication, motion planning & Object recognition, object detection  \\ \hline
		Agriculture: crop sensing; pest control (spraying and dusting).
		& Localization, motion control, coverage path following/trajectory tracking, motion planning, object detection & Object recognition, communication  \\ \hline
		Wildlife monitoring: tracking of poachers; animal census; animal behaviour monitoring.
		& Localization, motion control, path following/trajectory tracking, communication, motion planning, object detection & Object recognition, object tracking  \\ \hline
		Law enforcement: surveillance and monitoring; tracking of subjects; public address system; forensics.
		& Localization, motion control, path following/trajectory tracking, communication, object detection & Object recognition, object tracking  \\ \hline
	\end{tabularx}
	\end{center}
\end{table*}

\subsection*{Reliability}

\noindent Reliability is one of the two measures of belief in a system's ability to satisfactorily perform a particular task, the other being responsiveness. It is ascertained through compliance with the established performance requirements. A similarity function applied to the actual and required performance of a capability constitutes the acceptance policy. Since the adopted measure of reliability is variance, the similarity function returns a quotient of required to actual uncertainty, equivalently the quotient of actual to required precision. This formulation is beneficial in a way that it eliminates the subjective urge to compare different capabilities with the aim of determining one as better. This changes the focus of reliability assessment from the capability itself to its similarity to a required performance. An example of this would be lane following by an autonomous vehicle. This task can be made more challenging by requiring the vehicle to track the lane to within \(\pm4\%\) of the lane width or less challenging by requiring the vehicle to track the lane to within \(\pm20\%\) of the lane width. It should be noted that in doing so, the underlying lane tracking technology enabling achievement of the set requirement is irrelevant. \par

Mathematically, reliability \(C_{rel,i}\) for a capability \(i\), is expressed as in Eq.~\ref{eqn_TF_with_specialCases}, where \(\sigma_{ref}^2\) and \(\sigma_{act}^2\) are the variances of the Gaussian distributions related to the required and actual system errors, respectively. The former could be set by the manufacturer or regulatory authorities, while the latter is determined empirically for a particular task and operating domain. Despite location-invariance, variance is not linear in scale. But by taking a quotient of two variances, the scale non-linearity is normalised. The essential performance requirement \(\sigma_{ref, i}\) is selected in such a way as to reflect the criticality of the task and operating environment. The resulting reliability is unitless and non-negative, i.e,  \(C_{rel, i} \in \left[0, \infty\right)\).

\begin{equation} \label{eqn_TF_with_specialCases}
	{C_{rel,i}} =
	\begin{cases}
		{\frac{\sigma^2_{ref,i}}{\sigma^2_{act,i}},} & {\text{for } \sigma^2_{ref,i} \ge \sigma^2_{act,i} > 0},\\
		{0,} & {\text{for } \sigma^2_{act,i} > \sigma^2_{ref,i} \ge 0},\\
		{\infty,} & {\text{for } \sigma^2_{act,i} = 0}
	\end{cases}
\end{equation}

Experiments that involve observations of the probability of success \(p\) are modelled as Binomial distributions with variance calculated as \(\sigma^2=np(1-p)\), where \(n\) is the number of independent trials. \par 

\begin{figure}[!ht]
	\centering
	\includegraphics[width=0.9\linewidth]{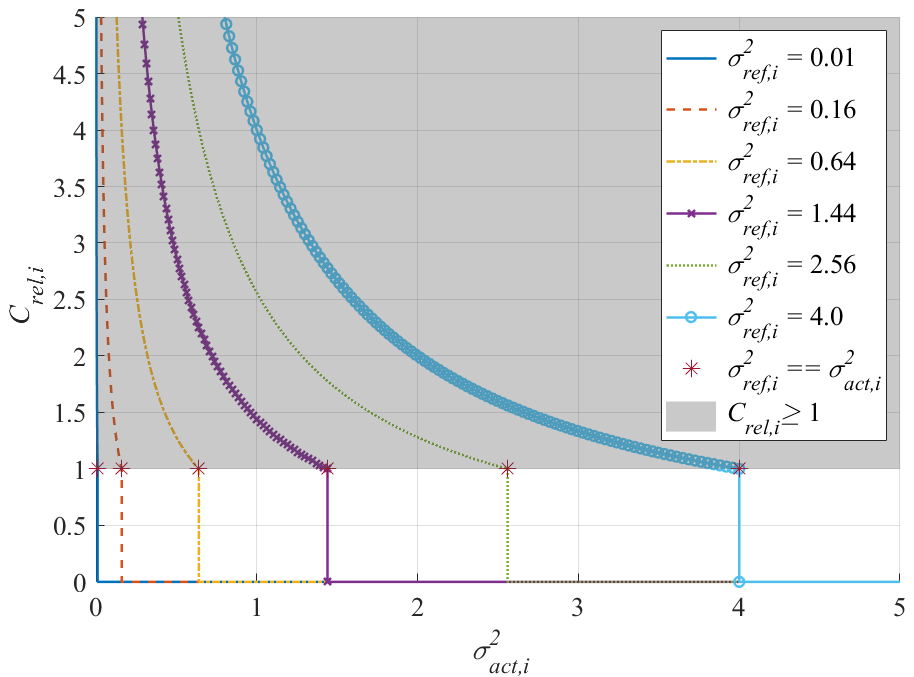}
	\caption{Plot of \(C_{rel,i}\) for several \(\sigma_{ref}^2\) values.}
	\label{fig_C_rel}
\end{figure}

Figure~\ref{fig_C_rel} shows plots of \(C_{rel}\) for several \(\sigma_{ref}^2\) values. A capability \(i\) is autonomously functional with respect to reliability if \(C_{rel,i} \ge 1\), which is a necessary, but not sufficient condition for unconditional full autonomous functioning with respect to reliability. 

\begin{figure}[!ht]
	\centering
	\includegraphics[width=0.8\linewidth]{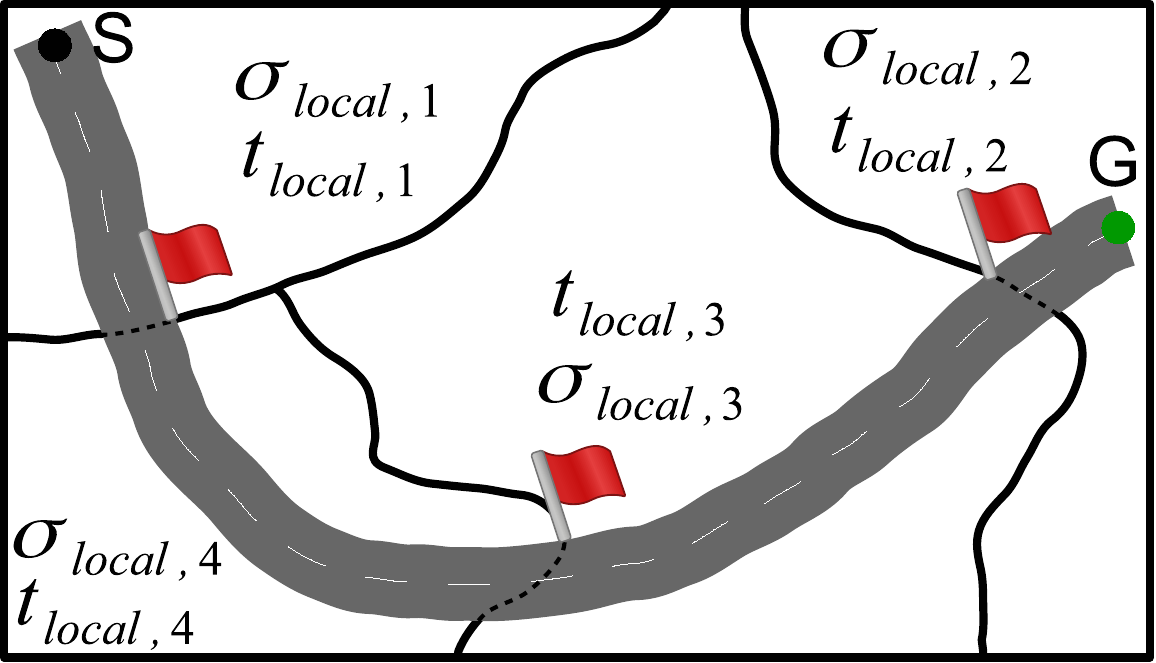}
	\caption{Operating environment with regions of different functional performance requirements for lateral vehicle motion control.}
	\label{fig_variance_map}
\end{figure}

As a demonstration, imagine the task of autonomous driving, where the environment is divided into operating regions each with a specific functional performance requirement for lateral vehicle motion control as indicated in figure.~\ref{fig_variance_map}. Then, if \(\sigma^2_{ref,i} \le \sigma^2_{global,i}\), where \(\sigma^2_{global,i} = \min \{\sigma^2_{local,1}, ... , \sigma^2_{local,j}, ... , \sigma^2_{local,k}\}\), fulfilling this requirement indicates unconditional fully autonomous functioning ability for capability \(i\) in all the \(k\) operating regions. But if \(\sigma^2_{ref,i} > \sigma^2_{global,i}\), then the associated full autonomous functioning ability is conditioned on reliability and only locally functional in regions where \(\sigma^2_{ref,i} \le \sigma^2_{local,j}\).

\subsection*{Responsiveness}

\noindent A capability can be developed to varying degrees of responsiveness, where responsiveness is parametrized by response time. An approach similar to that of reliability is taken in deriving the responsiveness metric. To achieve regulation, a responsiveness functional performance requirement is established and then the actual responsiveness is conditioned on this requirement using a similarity function. The mathematical representation of this metric is as indicated in Eq.~\ref{eqn_performanceCapacity}.

\begin{equation} \label{eqn_performanceCapacity}
	{C_{res,i}} =
	\begin{cases}
		\frac{t_{ref,i}}{t_{act,i}}, & \text{for } t_{ref,i} \ge t_{act,i} > 0 \\
		0, & \text{for } t_{act,i} > t_{ref,i} \ge 0
	\end{cases}
\end{equation} where \(t_{act,i}\) and \(t_{ref,i}\) are the actual and essential response times of capability \(i\). The resulting responsiveness is unitless and non-negative, \(0 \le C_{res,i} \le n\cdot t_{ref,i}\cdot\alpha\), where \(n\) is the number of operations per cycle and \(\alpha\) is the Bremermann's limit.
Note that, capability execution is an intrinsic characteristic of a capability and is conserved. \par

Some changes in the environment depend on the response time of the robotic system, but others are independent of this time. Therefore, the essential response time has to be selected so that its associated temporal frequency is greater than that of the environment or operating domain. \par

\begin{figure}[!ht]
	\centering
	\includegraphics[width=1.0\linewidth]{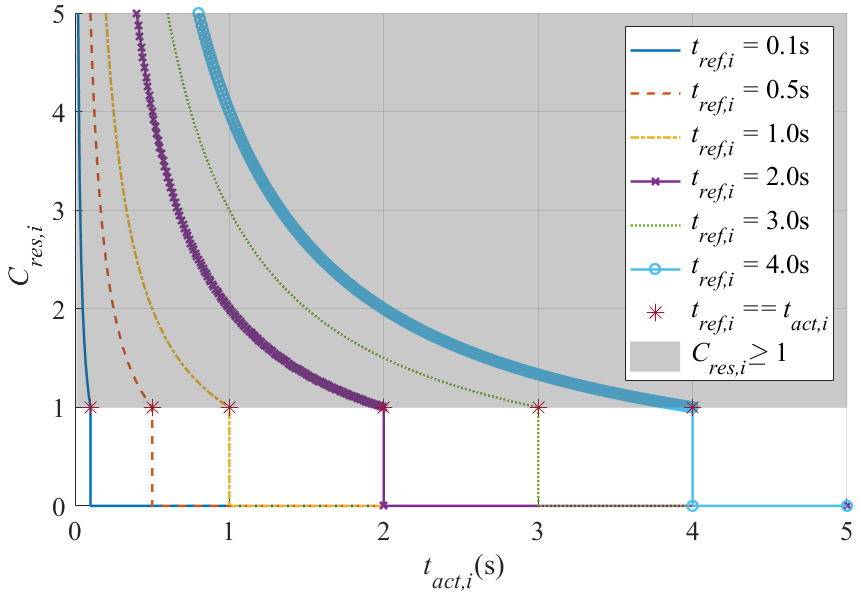}
	\caption{Plot of \(C_{res,i}\) for several \(t_{ref,i}\) values.}
	\label{fig_C_res}
\end{figure}

Figure~\ref{fig_C_res} shows plots of \(C_{res}\) for several \(t_{ref}\) values. A capability \(i\) is autonomously functional with respect to responsiveness if \(C_{res,i} \ge 1\), which is a necessary, but not sufficient condition for unconditional fully autonomous functioning with respect to responsiveness. \par

Let us extend the autonomous driving task in the previous sub-section by imposing response time functional requirements on each of the operating regions as indicated in figure~\ref{fig_variance_map}. Then, if \(t_{ref,i} \le t_{global,i}\), where \(t_{global,i} = \min \{ t_{local,1}, ..., t_{local,j}, ... , t_{local,k}\}\), fulfilling this requirement indicates unconditional full autonomous functioning ability for that capability in all the \(k\) operating regions. But if \(t_{ref,i} > t_{global,i}\), then the associated fully autonomous functioning ability is conditioned on responsiveness and only locally functional in regions where \(t_{ref,i} \le t_{local,j}\). \par

The proposed metrics are formulated in such a way as to not only provide a means of establishing measurable autonomy design goals, but also of regulating and ensuring autonomous functioning safety. The latter two are achieved through compliance with requirement specifications and integrity monitoring respectively. In the next section, the two measures of autonomy, namely level and degree of autonomy are introduced and described. 

\subsection{Level of Autonomy and Degree of Autonomy}
\label{sect_LoADoA}

\begin{figure}[!ht]
	\centering
	\includegraphics[width=1.0\linewidth]{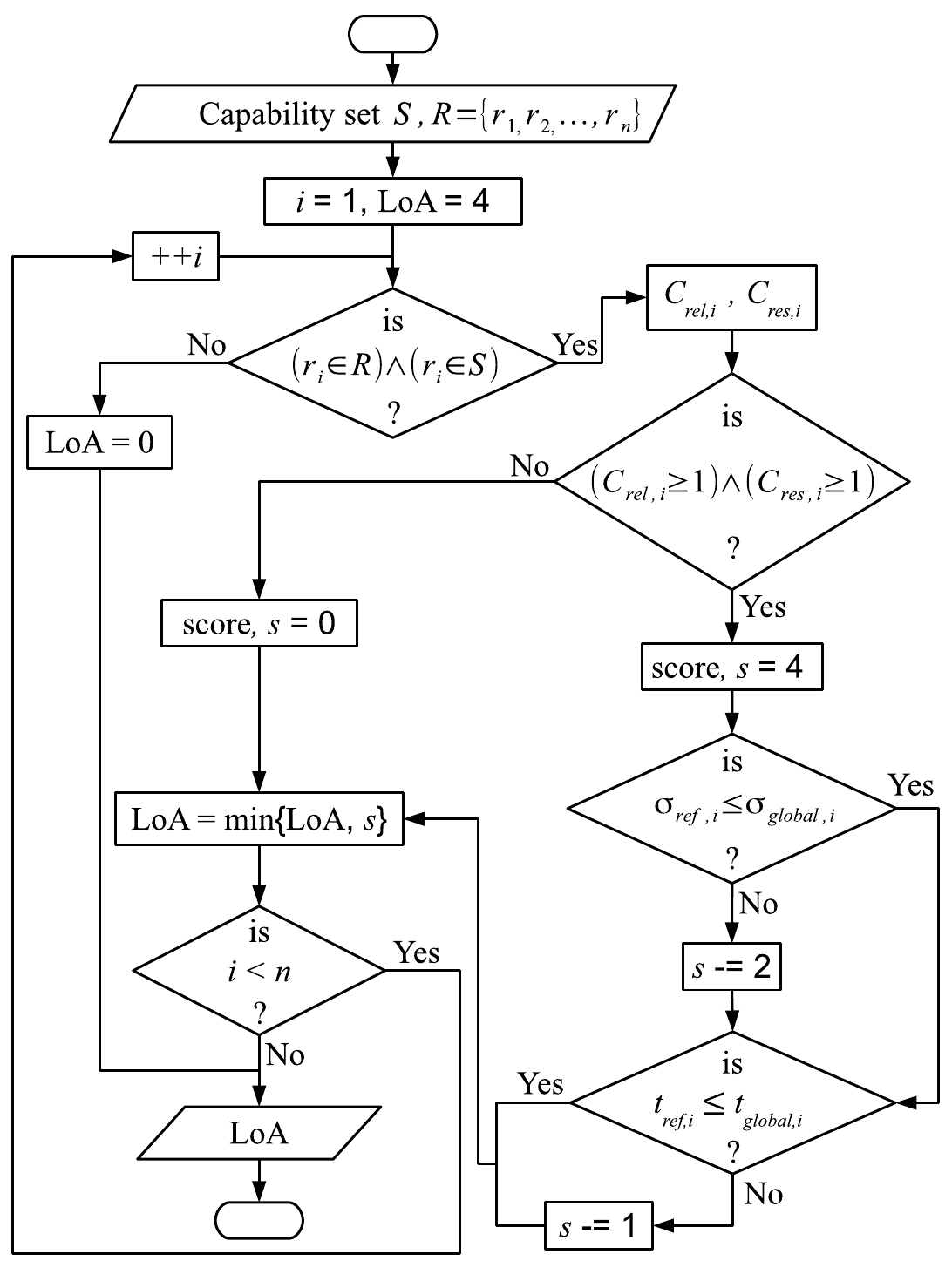}
	\caption{Proof of essential performance process.}
	\label{fig_AL_flowchart}
\end{figure}

\noindent Although neither reliability nor responsiveness is a sufficient condition for fully autonomous functioning, a logical combination of the two, i.e., \( C_{rel,i} \ge 1 \wedge C_{res,i} \ge 1\) is true, is sufficient, and is the primary test for LoA (Level of Autonomy). The LoA assessment procedure takes as input a requisite capability set \(R\) for a particular task, whose entries include the essential performance requirements for each of the \(n\) capabilities, and the actual system capability set \(S\). Then the entries go through a proof of essential performance process in figure~\ref{fig_AL_flowchart}, the output of which is an index corresponding to the system's LoA.

To describe the levels of full autonomy, this work applies the five-level chart in Table~\ref{tab_LFA}, which is derived from the ten-level chart in Table~\ref{tab_LAF} by grouping the ten levels into two categories, namely a category that does and a category that does not support external control. It is the latter category that none of the existing frameworks represent to such resolution as presented herein.  Each higher level in the former category represents an increase in machine control and a decrease in external control. As the focus is on fully autonomous functioning, all levels that support external intervention in Table~\ref{tab_LAF} are consolidated into level 0 in Table~\ref{tab_LFA}. \par

\renewcommand{\arraystretch}{1.3}
\linespread{1}
\begin{table}
	\begin{center}
	\caption{Levels of full autonomy in relevant environment.}
	\label{tab_LFA}
	\begin{tabularx}{0.95\linewidth}{c|X}\hline
		\textbf{Level} &  \textbf{Description} 	   \\ \hline\hline
		4 	 		   & Unconditional full autonomy. 	\\ \hline
		3 	 		   & Responsiveness-conditioned full autonomy.		\\ \hline
		2              & Reliability-conditioned full autonomy. 	\\ \hline
		1              & Responsiveness- \& reliability-conditioned full autonomy.   \\ \hline\hline
		0              & Externally controlled or supervised autonomous functioning. \\ \hline
	\end{tabularx}
	\end{center}
\end{table}

\renewcommand{\arraystretch}{1.3}
\linespread{1}
\begin{table}
	\begin{center}
	\caption{Level of autonomous functioning in relevant environment.}
	\label{tab_LAF}
	\centering
	\begin{tabularx}{0.95\linewidth} {
			 >{\centering\arraybackslash}c
			| >{\centering\arraybackslash}c
			| >{\raggedright\arraybackslash}X  }\hline
		\textbf{Level} & &  \textbf{Description} \\ \hline\hline
		9	 		   &\multirow{4}{*}{\rotatebox[origin=c]{90}{\parbox[c]{1.6cm}{\centering No external operator}}} & Unconditional full autonomy. 	\\ \cline{1-1}\cline{3-3}
		8 	 		   & & Responsiveness-conditioned full autonomy. 		\\ \cline{1-1}\cline{3-3}
		7              & & Reliability-conditioned full autonomy. 	\\ \cline{1-1}\cline{3-3}
		6              & & Responsiveness- \& reliability-conditioned full autonomy.    \\ \hline\hline
		5              &\multirow{6}{*}{\rotatebox[origin=c]{90}{\parbox[c]{3.5cm}{\centering External operator}}} & Full autonomous functioning with on-request supervision. \\ \cline{1-1}\cline{3-3}
		4              & & Full autonomous functioning with continuous supervision.		   \\ \cline{1-1}\cline{3-3}
		3              & & System proposes course of action, operator approves or modifies it for the system to execute.	   \\ \cline{1-1}\cline{3-3}
		2              & & System proposes course of action for the operator to approve or modify and execute.	   \\ \cline{1-1}\cline{3-3}
		1             & & Operator determines and provides course of action for the system to execute.	   \\ \cline{1-1}\cline{3-3} \noalign{\vskip\doublerulesep
			\vskip-\arrayrulewidth} \cline{1-1}\cline{3-3}
		0              & & Externally operated or remotely controlled. \\ \hline
	\end{tabularx}
	\end{center}
\end{table}

LoA considers the minimum performance requirement specifications to qualify a system's ability to accomplish a task autonomously in a particular operating domain, but one is also interested in knowing how high above this minimum the performance is. This is what DoA represents. Its mathematical formulation was inspired by the observation that performance of systems naturally gravitates towards the principle of increasing intelligence with decreasing precision \cite{saridis1989analytic}.  Therefore, by prescribing precision, the natural flow can be manipulated to yield the appropriate intelligence. This way, high levels of autonomy can be achieved without compromising on precision. \par

The proposed DoA model is as indicated in Eq.~\ref{eqn_doa}. It utilises reliability and responsiveness metrics to determine the performance of a system. It is analogous to average kinetic energy in a system since under the zero mean assumption, variance is proportional to the power of error\cite{ogata1995discrete}, it can also be interpreted as a quality factor.

\begin{equation} \label{eqn_doa}
	\text{DoA} = n^{2}\cdot \left[\sum\limits_{i=1}^n \frac{1}{C_{rel,i}\cdot C_{res,i}}\right]^{-1}
\end{equation}

The DoA model in Eq.~\ref{eqn_doa} assumes capabilities to be of equal importance, but this need not always be the case. When requisite capabilities are assigned different importance weights, the weighted DoA model in Eq.~\ref{eqn_weightedDoa} is applied.

\begin{equation} \label{eqn_weightedDoa}
	\text{DoA} =  n\cdot \left. \left[\sum\limits_{i=1}^n \left(\frac{w_{i}}{C_{rel,i}\times C_{res,i}}\right) \right/ {\sum\limits_{i=1}^n w_{i}} \right]^{-1}
\end{equation}

\begin{figure}[!ht]
	\centering
	\includegraphics[width=1.0\linewidth]{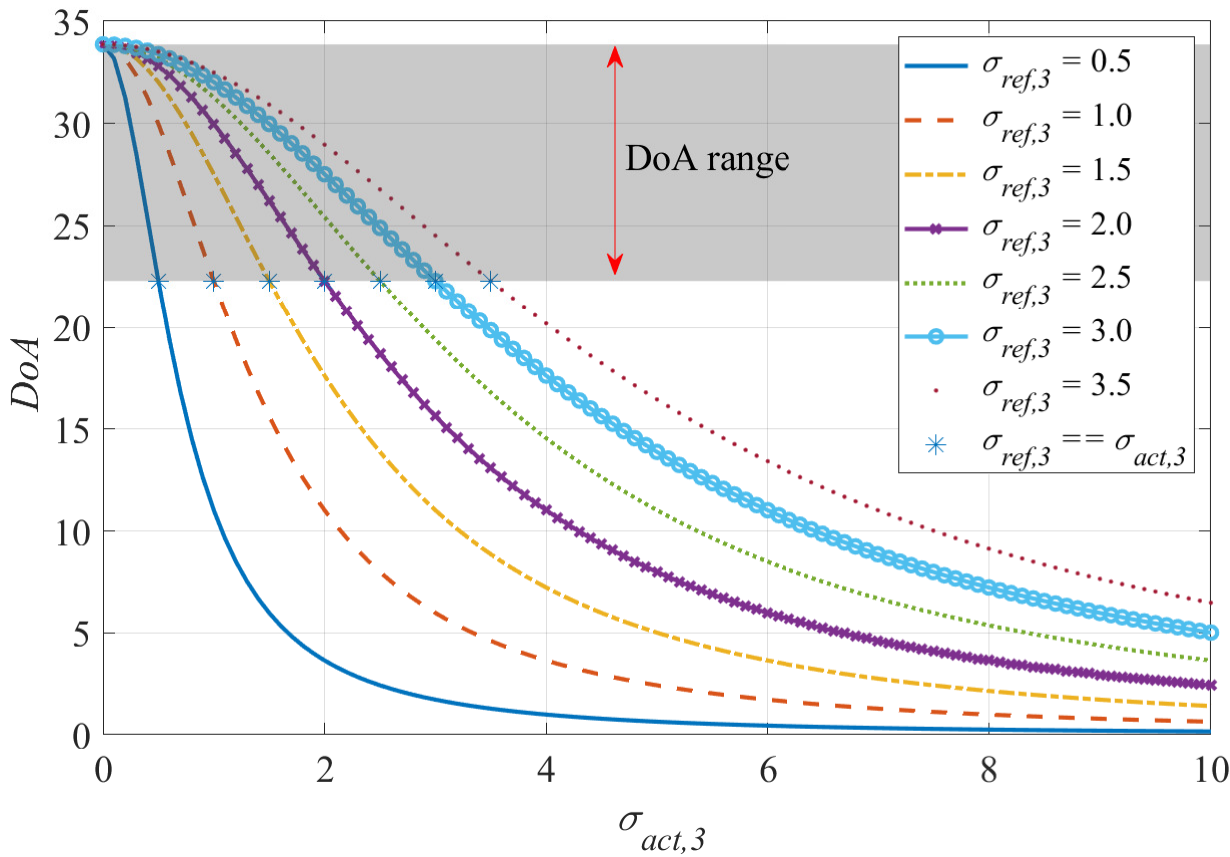}
	\caption{Varying the standard deviation of one capability for several reliability reference values. This figure was generated by varying the heading control capability of the autonomous driving case study in Sect.~\ref{sect_autonomousDriving}.}
	\label{fig_varyingCap_3}
\end{figure}

The characteristics of this DoA model are:
\begin{enumerate}
	\item{\(\text{DoA}\in[n,+\infty)\) for a requisite capability set of cardinality \(n\).}
	\item{The influence of a capability \(i\) is bounded, making it less sensitive to outlier performance.}
	\item{Matching reliability and responsiveness requirements of any capability \(i\) results in the same DoA contribution.}
	\item{The effect of varying the performance of a capability \(i\) is approximately linear around \(\sigma_{ref, i}\) and \(\nabla \text{DoA}(\sigma_{act, i}) \propto 1/\sigma_{ref,i}\). This is demonstrated in figure~\ref{fig_varyingCap_3}. Similar trends apply for response time.}
\end{enumerate}

\section{Integrity of Autonomous Functioning}
\label{sect_integrity}

\noindent The first step towards ensuring that autonomous systems perform in accordance with requirement specifications established by the industry or regulatory authorities is through product certification. As with other products, this certification should follow extensive system testing under all conceivable operating conditions over a specified period of time, which is not only a laborious and time consuming undertaking, but also past performance can not guarantee future performance for complex systems as their intrinsic properties change over time. This raises the need for online analysis to ensure regularity in performance, and this is the goal of integrity monitors. \par

Failure of any requisite capability constitutes a safety risk/hazard, which should be properly managed. Risk is evaluated as the product of the probability of failing and the severity of that failure. Hence, risk is the weighted severity. Assuming every failure to result in an accident and each accident to have the highest severity, then normalising the severity gives risk as just the probability of failure. \par

Given how often an autonomous capability should perform acceptably, the proposed integrity monitor tracks and detects when the performance becomes consistently unacceptable. Although the integrity of each requisite capability is monitored independently, together they determine the integrity associated with the overall task-fulfilling ability of the system. Therefore, robots are autonomous only to the extent that their performance is acceptable with respect to essential performance requirement specifications associated with the task at hand and the operating domain. The performance requirements are defined in terms of reliability, responsiveness and integrity risk. It should be noted that \(100\%\) system availability is implied, since the considered systems are fully autonomous. \par

Integrity is well studied in the aviation industry, specifically in relation with localization based on GNSS (Global Navigation Satellite System). In its basic form, integrity is parametrized by protection level (PL), alert limit (AL), time to alert (TTA) and integrity risk (IR) \cite{icao1996international}. Of the four, only PL and AL are directly related to the error, i.e., uncertainty and maximum acceptable error respectively, and their relationship is normally illustrated in the Stanford diagram. In the context of the proposed framework, actual and reference performances are analogous to PL and AL respectively. \par

A Failure event occurs not when the error exceeds the performance requirement, but when the statistical bound on actual performance falls below the reference performance. Furthermore, the actual error is normally not directly measurable, but its statistical bound (PL in this case) can be estimated using estimation algorithms. Unfortunately, linear and smug filters may exhibit divergence and overconfidences respectively, leading to unreliable error bound estimates and hence, safety issues. With strategic redundancy and establishment of a limit on occurrence of such events, loss of integrity can be detected. This limit is called the integrity risk, and defined as the probability that the integrity monitor fails to detect occurrence of a fault for a period longer than TTA \cite{annex2006vol}. \par

Adopting ASIL (Automotive Safety Integrity Level) of ISO 26262, the integrity risk ranges from \(1000\) to \(10\) dangerous failures per a billion operating hours, corresponding to least and highest severity hazards respectively. With the assumption of Gaussian distribution on error, the latter corresponds to approximately \(5.730729\sigma\), covering \(99.999999\%\) of errors. As a result, the alert limit is set to \(\textit{AL} = 5.730729\sigma_{ref}\) and protection level to \(\textit{PL} = 5.730729\sigma_{act}\). When the condition \(\frac{\textit{AL}}{\textit{PL}} = C_{rel} \ge 1\) is violated, an integrity event is triggered, indicating loss of integrity. This condition is integrated into both reliability and responsiveness metrics. Noteworthy is the temporal constancy assumption on capability execution period, which renders both \(t_{act}\) and \(t_{ref} \) constants for both offline and online scenarios, and hence do not contribute to integrity. \par

\section{Demonstration Case Studies}
\label{sect_caseStudy}

\noindent In this section, application demonstrations of the proposed framework are presented. The demonstrations include an autonomous dynamic driving task application and DARPA (Defense Advanced Research Projects Agency) subterranean (SubT) challenge rules analysis.

\subsection{Autonomous Driving}
\label{sect_autonomousDriving}
\noindent On-road vehicles are a good example because of the great need to eliminate human drivers. Despite the necessity, this goal is taking a gradual implementation with autonomous driving modules either cooperating or collaborating with human drivers, or operating under human oversight. But the increase in uncertainty that comes with support for human intervention in the control process of an autonomous vehicle necessitates the use of more sophisticated control strategies. Additionally, having a human in the loop may result in many variations of control architectures depending on the degree of human involvement in the control process, which makes it difficult to ensure reliability and assign liability in case of mishaps. Also, with driver intervention, it is difficult to tell how the vehicle would ultimately perform. To overcome these challenges, full autonomous driving is necessary. \par

Besides transportation, cars are designed with two additional goals, namely safety and comfort. But none of the autonomy metrics in use today, i.e., miles per disengagement, mean distance between interventions (MDBI) or mean time between interventions (MTBI) \cite{paz2020autonomous} address them. Secondly, these metrics are themselves targets and can be easily manipulated. In contrast, the proposed metrics comply with the Goodhart’s law rendering them good measures for the dynamic driving task. Their application follows two sequential steps, namely proof of capability check, i.e., LoA check and DoA analysis, i.e., functional performance analysis. \par

The requisite capabilities for a dynamic driving task include longitudinal and lateral position control, heading control, longitudinal and lateral speed control, object detection and local path planning. One can easily recognize these as the building blocks for driver-assistance technologies like lane keeping, adaptive cruise control, lane departure warning and park assist. Then, if the goal is autonomy assessment and regulation, all that is needed are autonomous driving performance requirements for vehicles to comply with. And like any other product, the only desiderata for allowance to participate in vehicular traffic is conformance with the requirements. \par

Consider a world with only two operating domains, namely motorways and built-up areas, their benchmark performance requirements for passenger service vans and two arbitrary vehicles A and B, and their performance specifications listed in Table~\ref{tab_DDT}. The requirements are determined from geometric characteristics of roadways, i.e., topology, curvature, width and topography. For ease of assignment, road segments with comparable geometric and topographic properties are governed by identical requirements.

\renewcommand{\arraystretch}{1.3}
\linespread{1}
\begin{table*}
	\begin{center}
	\begin{threeparttable}
		\caption{Dynamic driving task requirements for a passenger vehicle.}
		\label{tab_DDT}
		\begin{tabularx}{\linewidth} {
				>{\centering\arraybackslash}c
				| >{\raggedright\arraybackslash}p{3.4cm}
				| >{\centering\arraybackslash}X
				 >{\centering\arraybackslash}m{0.8cm}
				| >{\centering\arraybackslash}X
				 >{\centering\arraybackslash}m{0.8cm}
				| >{\centering\arraybackslash}m{0.8cm}
				|| >{\centering\arraybackslash}X
				 >{\centering\arraybackslash}m{0.8cm}
				| >{\centering\arraybackslash}X
				 >{\centering\arraybackslash}m{0.8cm}
				}\hline
			\multirow{2}{*}{\centering \(i\)} & \multirow{2}{*}{\centering \textbf{Capability}}  & \multicolumn{2}{c|}{\textbf{Motorways}} &  \multicolumn{2}{c|}{\textbf{Built-up Areas}} & \multirow{2}{1cm}{\centering \textbf{IR} (FPH)} & \multicolumn{2}{c|}{\textbf{A}} &  \multicolumn{2}{c}{\textbf{B}}\\ \cline{3-6} \cline{8-11}
			& & \(AL_{i}\) &  \(f_{ref, i}\) & \(AL_{i}\) &  \(f_{ref, i}\) & & \(PL_{i}\) &  \(f_{act, i}\) & \(PL_{i}\) &  \(f_{act, i}\) \\ \hline
			1 & Longitudinal position control & \qty{1.40}{\meter}\tnote{a} & \qty{150}{\hertz} & \qty{0.29}{\meter}\tnote{a} & \qty{150}{\hertz} & \(10^{-8}\) & \qty{1.00}{\meter} & 200Hz & \qty{0.15}{\meter} & \qty{160}{\hertz} \\ 
			2 & Lateral position control & \qty{0.57}{\meter}\tnote{a} & \qty{150}{\hertz} & \qty{0.29}{\meter}\tnote{a} & \qty{150}{\hertz} & \(10^{-8}\) & \qty{0.3}{\meter} & 200Hz & \qty{0.15}{\meter} & \qty{160}{\hertz} \\ 
			3 & Heading control & \qty{1.5}{\degree}\tnote{a} & \qty{150}{\hertz} & \qty{0.5}{\degree}\tnote{a} & \qty{150}{\hertz} & \(10^{-8}\) & \qty{1.0}{\degree} & \qty{200}{\hertz} & \qty{0.3}{\degree} & \qty{160}{\hertz} \\ 
			4 & Longitudinal speed control & \qty{4.1}{\kilo\meter\per\hour}\tnote{b} & \qty{150}{\hertz} & \qty{3.0}{\kilo\meter\per\hour}\tnote{b} & \qty{150}{\hertz} & \(10^{-8}\) & \qty{2.0}{\kilo\meter\per\hour}\tnote{b} & \qty{200}{\hertz} & \qty{2.0}{\kilo\meter\per\hour} & \qty{160}{\hertz} \\ 
			5 & Lateral speed control & \qty{1.0}{\kilo\meter\per\hour} & \qty{150}{\hertz} & \qty{1.4}{\kilo\meter\per\hour} & \qty{150}{\hertz} & \(10^{-8}\) & \qty{0.7}{\kilo\meter\per\hour} & \qty{200}{\hertz} & \qty{1.0}{\kilo\meter\per\hour} & \qty{160}{\hertz} \\ 
			6 & Object detection & \(95\%\) & \qty{10}{\hertz} & \(95\%\) & \qty{10}{\hertz} & \(10^{-7}\) & \(98\%\) & \qty{20}{\hertz} & \(96\%\) & \qty{15}{\hertz} \\ 
			7 & Local path planning & \(95\%\) & \qty{10}{\hertz} & \(95\%\) & \qty{10}{\hertz} & \(10^{-6}\) & \(99\%\) & \qty{20}{\hertz} & \(98\%\) & \qty{15}{\hertz} \\ \hline
		\end{tabularx}
		\footnotesize
		\begin{tablenotes}
			\item[a] Adopted from \cite{reid2019localization}. Based on lane width of \qty{3.6}{\meter} and top speed of \qty{137}{\kilo \meter \per \hour} on motorways, and lane width of \qty{3}{\meter} and minimum road curvature of \qty{20}{\meter} or width of \qty{3.3}{\meter} and road curvature of \qty{10}{\meter} for built-up areas.
			\item[b] Tolerance of vehicular traffic speed sensors on German roads is \(3\%\) for speeds over \qty{100}{\kilo \meter \per \hour}.
		\end{tablenotes}
	\end{threeparttable}
\end{center}
\end{table*}

The first step in assessing autonomy, proof of capability check yielded LoA 2 for vehicle A and LoA 4 for vehicle B. This means both vehicles possess the essential requisite capability, but the autonomy of vehicle A is conditioned on reliability as its passive protection levels are insufficient for built-up areas. Vehicle B possesses unconditional autonomy in this two-region world. The associated DoA is \(27524500\times10^{-6}\) for vehicle A on the motorways, and \(23469700\times 10^{-6}\) and \(19229600\times 10^{-6}\) for vehicle B on the motorways and built-up areas respectively. Although vehicle A does not meet all autonomous requirement for built-up areas, it outscores vehicle B on motorways. Depending on the performance of the seven requisite capabilities, DoA for autonomous dynamic driving in this world is in range \(7 \le \text{DoA} < +\infty\). \par

During operation, the system switches between open-loop and closed-loop phases, during computation and when computation is done respectively. For increased safety and decreased uncertainty, one should minimize the open-loop duration. Assuming the recommended maximum longitudinal speed of \qty{130}{\kilo\meter\per\hour} on German motorways and the longitudinal distance \(AL\) of \qty{1.4}{\meter} from Table~\ref{tab_DDT}. These correspond to an update rate of \qty{26}{\hertz}. With the goal of keeping the uncertainty due to computation delay as low as possible, the update rate is selected such that the resulting open-loop displacement is a fraction of the alarm limit. For example with \(5\times\)\qty{26}{\hertz}, the open-loop distance is \qty{0.28}{\meter}, while with \qty{150}{\hertz} and \qty{200}{\hertz} it is \qty{0.24}{\meter} and \qty{0.18}{\meter} respectively.

\subsection{DARPA Subterranean Challenge}
\label{sect_subT}

\noindent This section analyses the DARPA SubT challenge rules in light of the proposed metrics to show not only their prevalence, but also how their application can provide a self-evaluation measure indicative of a team's likelihood of qualifying for such competitions prior to qualification rounds. The SubT robotics competition consisted of two categories, namely systems and virtual competitions, aimed at motivating development of state-of-the-art solutions to navigation, mapping  and search tasks in dynamic, complex and unknown subterranean operating environments \cite{DefenseAdvancedResearchProjectsAgency2021}. In accomplishing those tasks, the competing systems utilized different sets of capabilities as indicated in Table~\ref{tab_subT_tasks}.  \par

\renewcommand{\arraystretch}{1.3}
\linespread{1}
\begin{table}
	\begin{center}
	\begin{threeparttable}
		\caption{Requisite capability sets for DARPA SubT challenge tasks.}
		\label{tab_subT_tasks}
		\begin{tabularx}{\linewidth} {
				 >{\raggedright\arraybackslash}p{1.9cm}
				| >{\centering\arraybackslash}p{1.1cm}
				 >{\centering\arraybackslash}X
				 >{\centering\arraybackslash}X
				| >{\centering\arraybackslash}p{0.65cm}
				 >{\centering\arraybackslash}p{0.65cm} }\hline
			\multirow{2}{2cm}{\textbf{Capability}} & \multicolumn{3}{c|}{\textbf{Tasks}} & \multicolumn{2}{c}{\textbf{Requirements}}\\ \cline{2-6}
			& Navigation & Mapping & Searching & \(AL_{i}\) & \(t_{ref,i}\) \\ \hline
			Localization & \checkmark & \checkmark & \checkmark & \qty{5}{\meter} & \textit{n.a} \\ \hline
			Motion planning & \checkmark & \checkmark & \checkmark & \textit{n.a} & \textit{n.a} \\ \hline
			Motion control & \checkmark & \checkmark & \checkmark & \textit{n.a} & \textit{n.a} \\ \hline
			Mapping  &  & \checkmark & & \textit{n.a} & \qty{10}{\second}\tnote{a} \\ \hline
			Communication  &  & \checkmark & \checkmark & \textit{n.a} & \textit{n.a} \\ \hline
			Path tracking & \checkmark & \checkmark & \checkmark & \textit{n.a} & \textit{n.a} \\ \hline
			\multirow{2}{*}{Object detection} & \multirow{2}{*}{} & \multirow{2}{*}{} & \multirow{2}{*}{\checkmark} & \(0.160\)\tnote{b} & \textit{n.a} \\ \cline{5-6}
			&  &  & & \(0.099\)\tnote{c} & \textit{n.a} \\ \hline
		\end{tabularx}
		\footnotesize
		\begin{tablenotes}
			\item[a] Communication delay assumed insignificant compared to mapping time.
			\item[b] Uncertainty for virtual competition.
			\item[c] Uncertainty for systems competition.
		\end{tablenotes}
	\end{threeparttable}
	\end{center}
\end{table}

The scoring objective for the final competition was the total number of accurately identified and localized artefacts within \qty{60}{\minute}. To be valid, the position of the artefact had to be within \(\pm\)\qty{5}{\meter} of the true position. Since the artefacts were spatially distributed within the environment, and the time limited, this implicitly bounded the exploration speed. The systems competition was evaluated based on one final run, which did not account for variability in performance, whereas in the virtual competition, the final score was averaged on \(m\) scenarios and \(n\) trial runs per scenario to account for random variability in performance. \par

With the aim of reporting all artefacts, it is observed that both competitions allowed five false reports, resulting in a combined uncertainty of correct identification and localization of an artefact of \(0.099\) and \(0.160\) for systems and virtual competitions respectively. Another indirectly specified requirement is the map update rate inferrable from the frequency of the mapping and map transfer process, which is \qty{0.1}{\hertz} at the most, and localization tolerance inferrable from the \qty{5}{\meter} artefact localization standard deviation. Nevertheless, ambiguities in the rules resulting from compounded performance requirements offered flexibility to developers as several lower-level performance combinations could meet the desired higher-level performance. This flexibility came at a cost, as regulation at the capability level was lost. This could have jeopardized the safety of the environment as a consequence of developers doing anything within their means to achieve the requirements.\par

From the right-hand side of Table~\ref{tab_subT_tasks}, it is evident that most of the requirements were left for the developers to specify, a similar stance has been taken by regulators of the autonomous driving industry. Therefore, developers of successful competing systems had to conduct additional requirements elicitation to guide the design specification of their vehicles. \par

\section{Conclusion}
\label{sect_conclusion}

\noindent Assessing autonomy is complicated by lack of clear terminology and metrics, two problems addressed by the energy-based framework presented herein. The framework is based on three quantitative metrics, namely requisite capability set, reliability and responsiveness, which were derived from a relationship between human-job and robot-task characteristics. With the metrics as inputs, the framework outputs a two-part measure of autonomy of a fully autonomous system, namely an interval-scale LoA and a ratio-scale DoA measure. LoA  uses an autonomy level chart to assess the existence and compliance of requisite capabilities with essential performance requirements, while DoA uses an average kinetic energy analogous model to assess the functional performance at solving a task. \par

The framework has been demonstrated on an on-road dynamic driving task, where it has been showed to exhibit the ability to limit the influence of outlier performances and linear behaviour around reference requirements. But most important is the fact that its mechanism of conditioning actual performance on required performance provides a regulatory tool for autonomous driving that can be applied by the industry and national departments of motor vehicles. \par

To ascertain the prevalence and relevance of the proposed metrics, we have showcased their extraction from the competition rules of the concluded DARPA subT challenge. Besides being easily obtainable, the performance specifications of these metrics fall in a continuous range \([0,\infty)\). Hence satisfy the three desired characteristics of autonomy metrics namely easily measurable, broad enough to capture autonomy evolution and with good output resolution. \par

The ubiquity of the proposed metrics in engineering applications is unquestionable as the inaccessibility to actual error has most performance measures directly or indirectly associated with uncertainty. This application of uncertainty measure addresses the "what-ifs" surrounding autonomy, for example, what if one system has more sensors? what if one system has more accurate sensors? what if one system has faster sensors? what if one system has a better state estimator? what if one system has a faster processor? what if one system is using a data-driven model? \par

Performance specifications can be established as standards that autonomous systems must operate in accordance with and could have either local or global validity. However, it is up to the individual system developers to devise their own approaches for achieving these directives. This approach has been inspired by the control and cybernetic loops. Therefore, it is our conviction that there is no need for an autonomy framework to extend its analysis to lower-level hardware or software specifications, where a lot of variations and little regulation may be exercised. With standardization, we envision a boost in user acceptance as they develop reliable expectations of autonomous performance. \par

As safety is an integral part of autonomy, higher levels of autonomy should be associated with higher levels of safety. As a results, herein an integrity monitoring system based on protection level and alert limit has been proposed. By tracking capability integrity online, the system’s contextual autonomy can be adapted accordingly or reasoned upon to engage the associated fail-safe mode or to decide on the right time for on-request supervision in systems that support external intervention. Such a monitoring system is capable of detecting faults including insufficient or excessive actuation, loss of control of a capability and destructive interference of capabilities, but cannot address failure of the whole system and anomaly elimination. \par

The framework not only assesses, but also facilitates the process of developing functional autonomous capability. The former streamlines and hence simplifies the process of regulating autonomous functioning. Owing to the DoA measure, the framework is capable of quantitatively ranking systems at a similar LoA on the basis of their performance. It is also extendable to multi-agent applications as only a clear definition of capability and its requirements are needed.\par

Although impactful, the framework makes a number of assumptions, namely capabilities are orthogonal and execute concurrently, knowledge of the environment/operating domain specifications, enforces a Gaussian distribution on performance and \(100\%\) availability of autonomous functionality, which may be impractical in some applications.  \par

Follow-up work will include extension of the integrity monitor to include availability and continuity for improved reliability, and lifting of the temporal constancy assumption on capability execution time. Then a benchmark study will be conducted for some of the aerial and ground autonomous vehicles in use today.

\balance
\bibliographystyle{IEEEtran}
\bibliography{References}

\end{document}